\documentclass[review]{elsarticle}

\usepackage{lineno,hyperref,subcaption}
\usepackage{algorithm}
\usepackage{amsmath}
\usepackage{amsfonts}
\usepackage{amssymb}
\usepackage[noend]{algpseudocode}
\modulolinenumbers[5]

\journal{}

\begin{document}

\begin{frontmatter}

\title{A Non-Rigid Map Fusion-Based RGB-Depth SLAM Method for Endoscopic Capsule Robots}


\author[address1,address2]{Mehmet Turan}
\ead{turan@is.mpg.de}

\author[address1,address2]{Yasin Almalioglu}
\ead{yasin.almalioglu@boun.edu.tr}

\author[address3]{Helder Araujo}
\ead{helder@isr.uc.pt}

\author[address2]{Ender Konukoglu}
\ead{ender.konukoglu@vision.ee.ethz.ch}

\author[address1]{Metin Sitti}
\ead{sitti@is.mpg.de}


\address[address1]{Max Planck Institute for Intelligent Systems, Stuttgart, Germany}
\address[address2]{Computer Vision Laboratory, ETH Zurich, Switzerland}
\address[address3]{Robotics Laboratory, University of Coimbra, Portugal}

\begin{abstract}
In the gastrointestinal (GI) tract endoscopy field, ingestible wireless capsule endoscopy is considered as a minimally invasive novel diagnostic technology to inspect the entire GI tract and to diagnose various diseases and pathologies. Since the development of this technology, medical device companies and many groups have made significant progress to turn such passive capsule endoscopes into robotic active capsule endoscopes to achieve almost all functions of current active flexible endoscopes. However, the use of robotic capsule endoscopy still has some challenges. One such challenge is the precise localization of such active devices in 3D world, which is essential for a precise three-dimensional (3D) mapping of the inner organ. A reliable 3D map of the explored inner organ could assist the doctors to make more intuitive and correct diagnosis. In this paper, we propose to our knowledge for the first time in literature a visual simultaneous localization and mapping (SLAM) method specifically developed for endoscopic capsule robots. The proposed RGB-Depth SLAM method is capable of capturing comprehensive dense globally consistent surfel-based maps of the inner organs explored by an endoscopic capsule robot in real time. This is achieved by using dense frame-to-model camera tracking and windowed surfel-based fusion coupled with frequent model refinement through non-rigid surface deformations.

\end{abstract}


\end{frontmatter}


\section{Introduction}

In the past decade, the advances in microsensors and microelectronics have enabled small, low cost devices in a variety of high impact applications. Following these advances, untethered pill-size, swallowable capsule endoscopes with an on-board camera and wireless image transmission device have been developed and used in hospitals for screening the gastrointestinal (GI) tract and diagnosing diseases such as the inflammatory bowel disease, the ulcerative colitis, and the colorectal cancer. Unlike standard endoscopy, endoscopic capsule robots are non-invasive, painless, and more appropriate to be employed for long-duration screening purposes. Moreover, they can access difficult body parts that were not possible to reach before with standard endoscopy (e.g., small intestines). Such advantages make pill-size capsule endoscopes a significant alternative screening method over standard endoscopy \citep{liao2010indications, nakamura2008capsule, pan2011swallowable, than2012review}.

However, current capsule endoscopes used in hospitals are passive devices controlled by peristaltic motions of the inner organs. The control over the capsule's position, orientation, and functions would give the doctor a more precise reachability of targeted body parts and more intuitive and correct diagnosis opportunity.  Therefore, several groups have recently proposed active, remotely controllable robotic capsule endoscope prototypes equipped with additional functionalities, such as local drug delivery, biopsy, and other medical functions \citep{sitti2015biomedical, yim2013magnetically, yim2012shape, yim2012design, goenka2014capsule, nakamura2008capsule, munoz2014review, carpi2011magnetically, keller2012method, mahoney2013managing, yim2014biopsy, petruska2013omnidirectional}. An active motion control is, on the other hand, heavily dependent on a precise and reliable real-time pose estimation capability, which makes the robot localization and mapping the key capability for a successful endoscopic capsule robot operation. In the last decade, several localization methods \citep{than2012review, fluckiger2007ultrasound, rubin2006sonographic, kim2008noninvasive, yim20133} were proposed to calculate the 3D position and orientation of the endoscopic capsule robot such as fluoroscopy \citep{than2012review}, ultrasonic imaging \citep{fluckiger2007ultrasound, rubin2006sonographic, kim2008noninvasive, yim20133}, positron emission tomography (PET) \citep{than2012review, yim20133}, magnetic resonance imaging (MRI) \citep{than2012review}, radio transmitter based techniques, and magnetic field-based techniques \citep{yim20133, son20165}. The common drawback of these localization methods is that they require extra sensors and hardware to be integrated to the robotic capsule system. Such extra sensors have their own drawbacks and limitations if it comes to their application in small-scale medical devices, such as space limitations, cost aspects, design incompatibilities, biocompatibility issue, accuracy, and the interference of the sensors with the activation system of the device. 

\section{Visual SLAM Structure and Survey}
As a solution of these issues, a trend of vision-based localization methods have attracted the attention for the localization of such small-scale medical devices. With their low cost and small size, cameras are frequently used in localization applications where weight and power consumption are limiting factors, such as in the case of small-scale robots. However, many challenges posed by the GI tract and low quality cameras of the endoscopic capsule robots are causing further difficulties in front of a vision based SLAM technique to be applied in that field. One of the most important challenges is the non-rigid structure of the GI tract, which causes deformations of the organ tissue making robotic mapping and localization even more difficult. Self-repetitiveness of the GI tract texture, heavy reflections caused by the organ fluids, peristaltic motions of the inner organs, and lack of distinctive feature points on the GI tract tissue are further challenges in front of a reliable robotic operation. Moreover, the low frame rate and resolution of the current capsule camera system also restrict the applicability of computer vision methods inside the GI tract. Especially visual localization methods based on feature point detection and tracking have less performance in the abdomen region than other application areas such as outdoor or indoor large scale environments.

Such a modern visual SLAM method is expected to be equipped with an accurate camera pose estimation module that is not affected by sudden movements, blur, noise, illumination changes, occlusions and large depth variations. Such a SLAM method should have a precise map reconstruction module that is capable of generating efficient dense scene representations in regions of little texture. A map maintenance method improving the map quality with resilience against dynamic changing small and large scale environments and a failure recovery procedure reviving the system from significantly large changes in camera viewpoints are further expected capabilities of such a modern SLAM system.


Modern visual SLAM methods are categorized as direct and indirect. Direct methods -also known as dense or semi-dense methods- exploit the information available at every pixel to estimate camera pose and to reconstruct the 3D map \citep{engel2014lsd}. Therefore, they are more persistent than indirect methods in regions with poor texture. Nevertheless, direct methods are susceptible to failure when scene illumination changes occur as the minimization of the photometric error assumes brightness consistency constraint between frames. 

 Indirect methods were introduced to reduce the computational complexity and achieve real time processing time. This is achieved by using sparse feature points instead all pixel information \citep{mur2015orb}. Features are expected to be distinctive and invariant to viewpoint and illumination changes, as well as resilient to blur, occlusion and noise. On the other hand, it is desirable for feature extractors to be computationally efficient and fast. Unfortunately, such objectives are hard to achieve at the same time and a trade-off between computational speed and feature quality is required.

Dense RGB Depth SLAM maps a space by fusing the RGB and depth data into a representation of the continuous surfaces it contains, permitting accurate viewpoint-invariant localization as well as offering the potential for detailed semantic scene understanding \citep{keller2012method}. However, existing dense SLAM methods suitable for incremental, real-time operation struggle when the sensor makes movements which are both of extended duration and often criss-cross loop back on themselves. Such a trajectory is typical of an endoscopic capsule robot aiming to explore and densely map the GI tract organs of a patient.

Despite there exists no work done for vision-based endoscopic capsule robot localization and mapping in the literature, some works exist for hand-held standard endoscopic SLAM. Visual SLAM for Hand-held Monocular Endoscope \citep{grasa2014visual} is one approach which uses visual features to localize hand-held monocular endoscopes. This paper claims to achieve good results but since the results were achieved in a very controlled environment with rigid and well textured surface and perfect lighting without reflections, the results are not comparable for challenge real scenes. ORB-SLAM-based Endoscope Tracking and 3D Reconstruction \citep{mahmoud2016orbslam} is another approach addressing the issue of endoscopic localization. This algorithm uses ORB descriptors as detected landmarks and using these descriptors the images are tracked in real time and the endoscope is localized using an already available 3D map information. As mentioned earlier, the nature of endoscopic images makes it rather difficult to maintain the tracking until operation executes.

\section{System Overview and Analysis}

\begin{figure}
\centering
  \includegraphics[width=\textwidth]{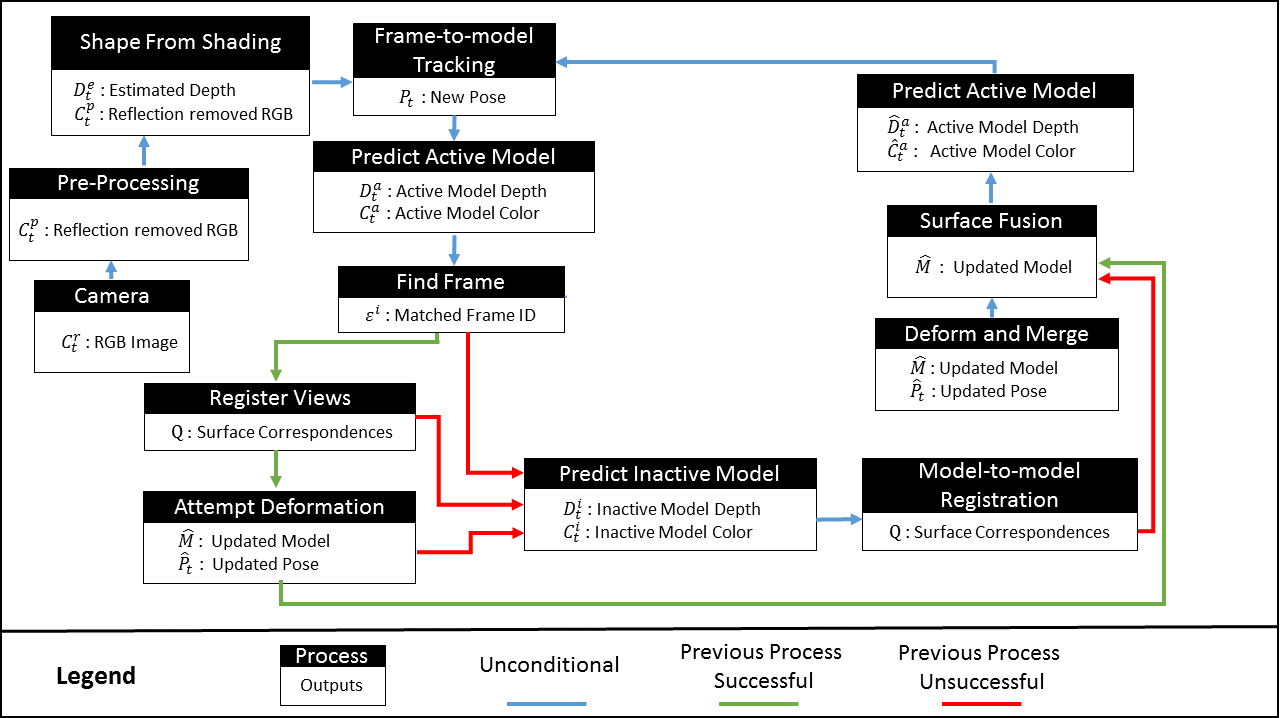}
\caption{System overview}
\label{fig:model_flowchart}       
\end{figure}

Our architecture follows the traditional approach in real-time dense visual SLAM systems that alternates between localization and mapping \citep{newcombe2011kinectfusion, whelan2015elasticfusion, keller2013real, henry2013patch, chen2013scalable, newcombe2011dtam, whelan2016elasticfusion}. Making significant use of GPU programming with CUDA library, robot localization module was implemented. OpenGL Shading Language was utilized to manage map reconstruction and maintenance. Our approach is based on estimating a fully dense 3D map of the inner organ tissue explored with an endoscopic capsule robot in real-time. In the following, key points of the proposed system are summarized (\ref{fig:model_flowchart}):

\begin{enumerate}
\item Detect and remove specular reflections using an authentic reflection suppression method;

\item Correct vignetting distortions;

\item Create 3D maps of consecutive endoscopic images using the Tsai-Shah surface reconstruction method \citep{ping1994shape};

\item Estimate a fused surfel-based model of the observed parts of the inner organ inspired by the surfel-based fusion system \citep{keller2013real}; 

\item Define the environment into active and inactive areas inspired by the frame-to-model fusion approach \citep{whelan2016elasticfusion}. Areas are getting inactive if they have not appeared for a certain period of time $\delta t$. Tracking and fusing modules use data in the active area and do not use inactive areas.

\item Every frame, search for intersecting parts of active and inactive areas of the model. In case there exists an intersection, fuse the portion of the active model within the current estimated camera frame with the portion of the inactive model in the same frame. A successful fusion realizes the loop-closure to the previously inactive model by deforming the entire model in a non-rigid fashion. The previous inactive part is reactivated and served for a tracking and surface fusion between the registered areas.

\end{enumerate}

At the time of writing, besides being the first visual SLAM method specifically developed for endoscopic capsule robots, we believe our proposed system to be the first of its kind for hand-held endoscopic SLAM methods with many proposed novelties such as the use of photometric and geometric frame-to-model predictive tracking in a fused surfel-based dense map, use of dense model-to-model local surface loop closures with a non-rigid space deformation.

\subsection{Preprocessing}
The proposed system starts with a preprocessing module that suppresses reflections caused by inner organ fluids. The reflection detection is done  by combining the gradient map of the input image with the peak values detected by an adaptive threshold operation. For the suppression of the detected reflections, inpainting method is applied onto the detected reflection distorted areas.

\subsection{Shapes from shading-based 3D map reconstruction}
\begin{figure}
	\centering
	\includegraphics[width=\textwidth]{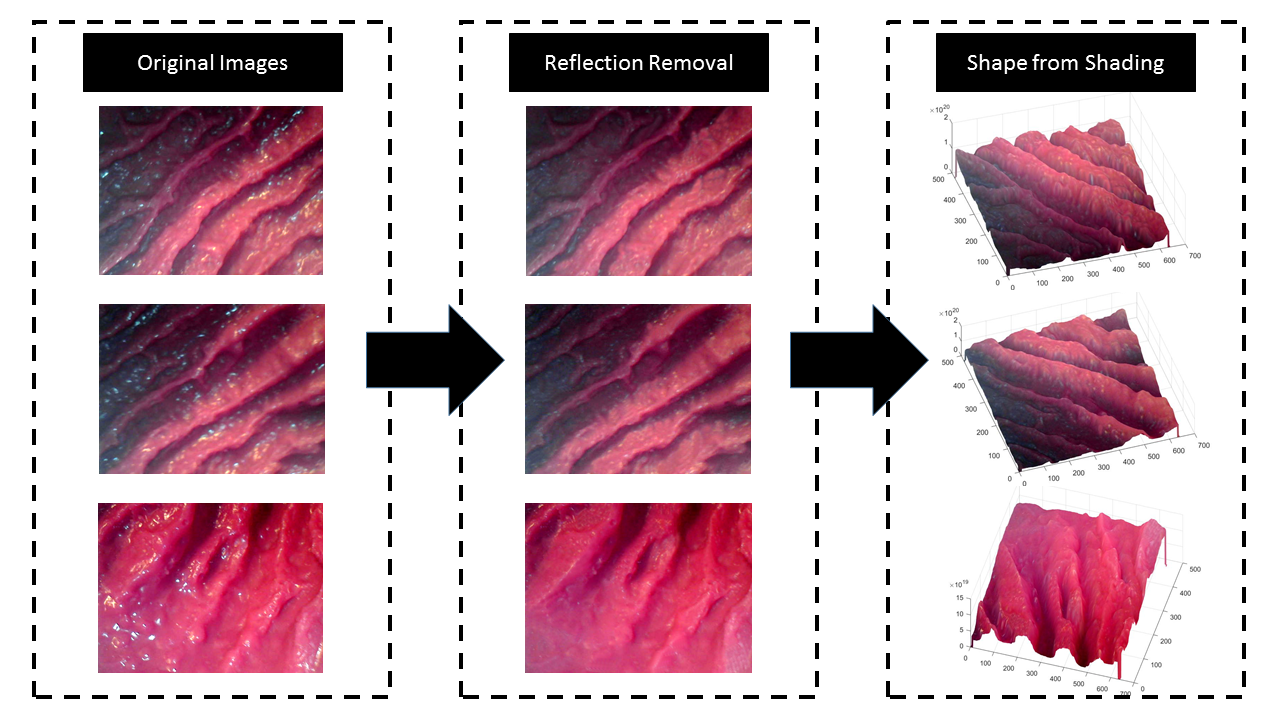}
	\caption{Reflection removal and shape from shading-based 3D map reconstruction.}
	\label{fig:sfs_new}       
\end{figure}

Since the light source direction is known both in endoscopic capsule robot applications and standard hand-held endoscope systems, shape from shading technique is an effective and powerful method to reconstruct the 3D surface of the GI tract. In our system, we implemented the Tsai and Shah surface reconstruction method \citep{ping1994shape}. This method uses a linear approximation to calculate depth map in an iterative manner using the estimated slant, tilt and albedo values. For further details of the method, the reader is suggested to refer to the original paper of the method. Figure \ref{fig:sfs_new} demonstrates the results after applying the reflection suppression and shape from shading-based 3D surface reconstruction module.

\subsection{Fused predicted tracking}
Our scene representation is an unordered list of surfels $\mathcal{M}$ inspired by the representation used in \citep{keller2012method}. Each surfel $\mathcal{M}^s$ has a position $\textbf{p} \in \mathbb{R}^3$, normal $\textbf{n}\in \mathbb{R}^3$, colour $\textbf{c}\in\mathbb{N}^3$, weight $w\in\mathbb{R}$, radius $r \in \mathbb{R}$, initialisation timestamp $t_0$ and last updated timestamp $t$. The radius of each surfel is intended to represent the local surface area around a given point while minimising visible holes \citep{salas2014dense}. Surfel initialisation and depth map fusion follow the same scheme and surfel colours follow the same moving average approach as in \citep{keller2012method}. Pose estimation is realized by estimating a full colour splatted rendering of the surfels and performing a  photometric frame-to-model tracking \citep{whelan2016elasticfusion}. A time window threshold $\delta t$ is predefined which labels the surfels in $\mathcal{M}$ as active and inactive. Only active surfels are taken into account for robot trajectory estimation and depth map fusion. A surfel in $\mathcal{M}$ is declared as inactive when the time since that surfel was last updated is greater than $\delta t$ \citep{whelan2016elasticfusion}.
\subsubsection{Joint photometric and geometric pose estimation from a splatted surfel prediction}

We define the image space domain as $\Omega\subset\mathbb{N}^2$, where an RGB-D frame is composed of a depth map $\mathcal{D}$ of depth pixels $d:\Omega\rightarrow \mathbb{R}$ acquired by shape from shading technique and a colour image $\mathcal{C}$ of colour pixels acquired by the RGB camera of the endoscopic capsule robot $\textbf{c}:\Omega\rightarrow\mathbb{N}^3$. A normal map is extracted for every depth map. Given a depth map $\mathcal{D}$, the 3D back-projection of a point $\textbf{u}\in\Omega$ is defined as $\textbf{p}(\textbf{u},\mathcal{D}) = \textbf{K}^{-1}\textbf{\textit{u}}d(\textbf{u})$, where  $\textbf{K}$ is the camera intrinsics matrix and $\textbf{\textit{u}}$ is the homogeneous form of $\textbf{u}$. The intensity value of a pixel $\textbf{u} \in \Omega$ given a colour image $\mathcal{C}$ with colour  $\textbf{c}(\textbf{u}) =[c_1,c_2,c_3]^T$ is defined as $I(\textbf{u},\mathcal{C}) = (0.2989\times c_1 + 0.5870\times c_2 + 0.1140\times c_3) $ as predefined NTSC standards. Current depth map and colour image will be fused with the surfel-splatted predicted depth map and colour image of the current active model to determine the global pose  $\textbf{P}_t$ of the camera. This is called joint photometric and geometric pose estimation approach. All camera poses are represented with a transformation matrix where
\\
\begin{equation}
\textbf{P}_t = \begin{bmatrix}
       & \textbf{R}_t & &\textbf{t}_t            \\[0.3em]
       0 & 0 & 0 & 1           
     \end{bmatrix}
     \in \mathbb{SE}_3.
\end{equation}


The geometric pose $\xi$ between the current depth map $\mathcal{D}^l_t$ and the predicted active depth map from the last frame $\hat{\mathcal{D}}_{t-1}^a$ is calculated by minimizing the cost function over the point-to-plane error between 3D back-projected vertices:
\begin{equation}
E_{icp} = \sum_k{((\textbf{v}^k - \exp{(\hat{\xi})}\textbf{T} \textbf{v}_k^t)\cdot\textbf{n}^k)^2}
\end{equation}
where $\textbf{v}^k_t$ is the back-projection of the $k$-th vertex in $\mathcal{D}^l_t$, $\textbf{v}^k$ and $\textbf{n}^k$ are the corresponding vertex and normal represented in the previous camera coordinate frame. $\textbf{T}$ is the current estimate of the transformation from the previous camera pose to the current one and $\exp{(\hat{\xi})}$ is the matrix
exponential that maps a member of the Lie algebra $\mathfrak{se}_3$ to a member of the corresponding Lie group $\mathbb{SE}_3$. Vertices are associated using projective data association \citep{newcombe2011dtam}.


The photometric pose $\xi$ between the current colour image $\mathcal{C}^l_t$ and the predicted active model colour from the last frame $\hat{\mathcal{C}}^a_{t-1}$ is calculated by minimising the cost function over the photometric error between pixels which is defined as the intensity difference: 
\begin{equation}
E_{rgb} = \sum_{\textbf{u} \in \Omega} \left( I(\textbf{u},\mathcal{C}^l_t) - I(\pi(\textbf{K} \exp(\hat{\xi}) \textbf{T} \textbf{p}(\textbf{u} , \mathcal{D}^l_t)), \hat{\mathcal{C}}_{t-1}^a) \right)^2
\end{equation}
where as above $\textbf{T}$ is the current estimate of the transformation from the previous camera pose to the current one. The joint photometric and geometric pose is defined by  the cost function:
\begin{equation}
E_\textrm{track} = E_\textrm{icp} + w_\textrm{rgb}E_\textrm{rgb},
\end{equation}
with $w_\textrm{rgb} = 0.1$ in line with related work \citep{henry2013patch, whelan2015real}. For the minimisation of this cost function, the Gauss-Newton non-linear least-squares method with a three-level coarse-to-fine pyramid scheme is applied. 

\subsection{Deformation graph}
Loop closures are carried out in the set of surfels $\mathcal{M}$ by non rigid deformations according to the surface constraints provided by the local loop closure method in order to ensure surface consistency in the 3D reconstructed non rigid organ surface. For that purpose, a space deformation approach based on the embedded deformation technique of \citep{sumner2007embedded} is applied.

A deformation graph consists of a set of nodes and edges distributed throughout the model to be deformed. Each node $\mathcal{G}^n$ has a time-stamp $\mathcal{G}^n_{t_0}$, a position $\mathcal{G}_g^n \in \mathbb{R}^3$ and set of neighboring nodes $\mathcal{N}(\mathcal{G}^n$). The directed edges of the graph are the neighbors of each node. A graph is connected up to a neighbor count $k$ such that $\forall n,|\mathcal{N}(\mathcal{G}^n)| = k$. Each node also stores an affine transformation in the form of a $3\times 3$ matrix $\mathcal{G}^n_{\textbf{R}}$  and a $3\times 1$ vector $\mathcal{G}_{\textbf{t}}^n$ , initialized by default to the identity and $(0,0,0)^T$ respectively. When deforming a surface, the $\mathcal{G}^n_{\textbf{R}}$ and $\mathcal{G}_{\textbf{t}}^n$  parameters of each node are optimized according to surface constraints.

In order to apply a deformation graph to the surface, each surfel $\mathcal{M}^s$ identifies a set of influencing nodes in the graph $\mathcal{I}(\mathcal{M}^s, \mathcal{G})$. The deformed position of a surfel is given by \citep{whelan2016elasticfusion}:
\begin{equation}
\hat{\mathcal{M}}^s_{\textbf{p}} = \phi(\mathcal{M}^s) = \sum_{n \in \mathcal{I}(\mathcal{M}^s, \mathcal{G})} w^n(\mathcal{M}^s) [\mathcal{G}^n_{\textbf{R}}(\mathcal{M}_{\textbf{p}}^s - \mathcal{G}_{\textbf{g}}^n) + \mathcal{G}_{\textbf{g}}^n + \mathcal{G}_{\textbf{t}}^n]
\end{equation}
while the deformed normal of a surfel is given by:
\begin{equation}
\hat{\mathcal{M}}^s_{\textbf{p}} = \sum_{n \in \mathcal{I}(\mathcal{M}^s, \mathcal{G})} w^n (\mathcal{M}^s)\mathcal{G}^{{n-1}^T}_{\textbf{R}} \mathcal{M}^s_{\textbf{n}},
\end{equation}
where $w^n (\mathcal{M}^s)$ is a scalar representing the influence node $\mathcal{G}^n$ has on surfel $\mathcal{M}^s$, summing to a total of $1$ when $n = k$:
\begin{equation}
w^n (\mathcal{M}^s) = (1 - ||\mathcal{M}^s_{\textbf{p}}-\mathcal{G}_{\textbf{g}}^n ||_2 / d_\textrm{max})^2.
\end{equation}
Here, $d_\textrm{max}$ is the Euclidean distance to the $k+1$-nearest node of $M^s$. For the construction, optimization and application of the deformation graph into the map, we followed the rules represented by \citep{whelan2016elasticfusion}.

\subsection{Loop closure}
To ensure local surface consistency throughout the map our system closes many small loops with the existing map as those areas are revisited. We fuse into the active area of the model while gradually labeling surfels that have not been seen in a period of time $\delta t$ as inactive. The inactive area of the map is not used for frame tracking and fusion until a loop is closed between the active model and inactive model, at which point the matched inactive area becomes active again. This has the advantage of continuous frame-to-model tracking and also model-to-model tracking which provides viewpoint-invariant local loop closures \citep{whelan2016elasticfusion}.

We divide the set of surfels in our map $\mathcal{M}$ into two disjoint active and inactive sets. A match between the active and inactive sets is tried to be established by registering the predicted surface renderings of the sets from the latest pose estimate. Our algorithm does not check global loop closures since the inner organ volumes are quite small.

Before the activation of the deformation functionality, the algorithm checks the final condition of the Gauss-Newton optimization used to align the two views. If a high quality alignment has been achieved, a set of surface constraints $\mathcal{Q}$ are generated and fed into the deformation graph to align active and inactive surfels.

\section{EXPERIMENTS AND RESULTS} \label{sec:experiements}

We evaluate the performance of our system both quantitatively and qualitatively in terms of trajectory estimation, surface reconstruction accuracy and computational performance. 

\subsection{Dataset, equipment and specifications} \label{sec:dataset_equip}
\begin{figure}
	\centering
	\includegraphics[width=1\textwidth]{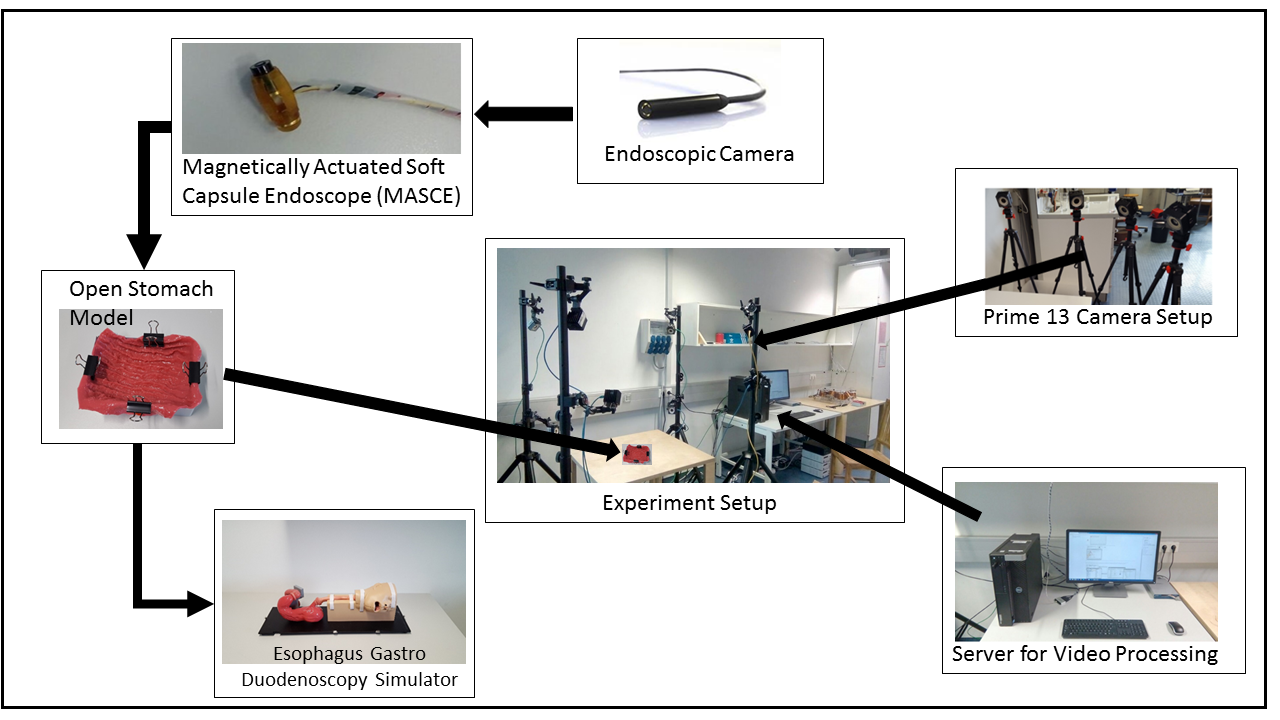}
	\caption{Experimental Setup}
	\label{fig:ex}       
\end{figure}

\begin{figure}
	\centering
	\includegraphics[width=0.9\textwidth]{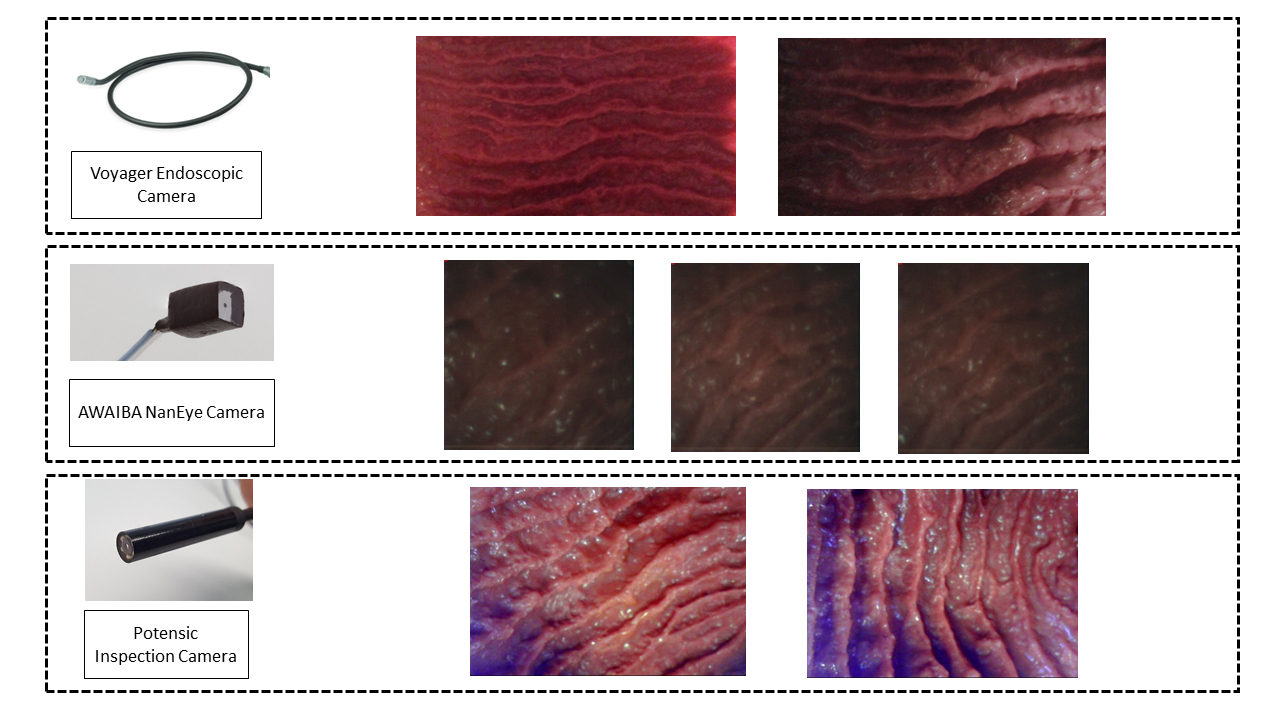}
	\caption{Sample images from our dataset}
	\label{fig:datas}       
\end{figure}

\begin{figure}
\centering
  \includegraphics[width=0.5\textwidth]{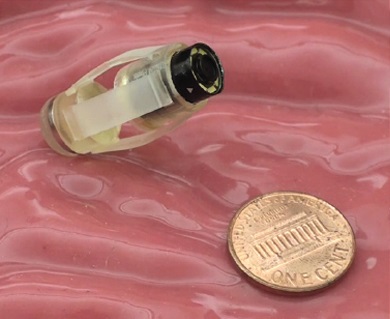}
\caption{Photo of the endoscopic capsule robot prototype used in the experiments.}
\label{fig:masce}       
\end{figure}

\begin{figure}
\centering
  \includegraphics[width=\textwidth]{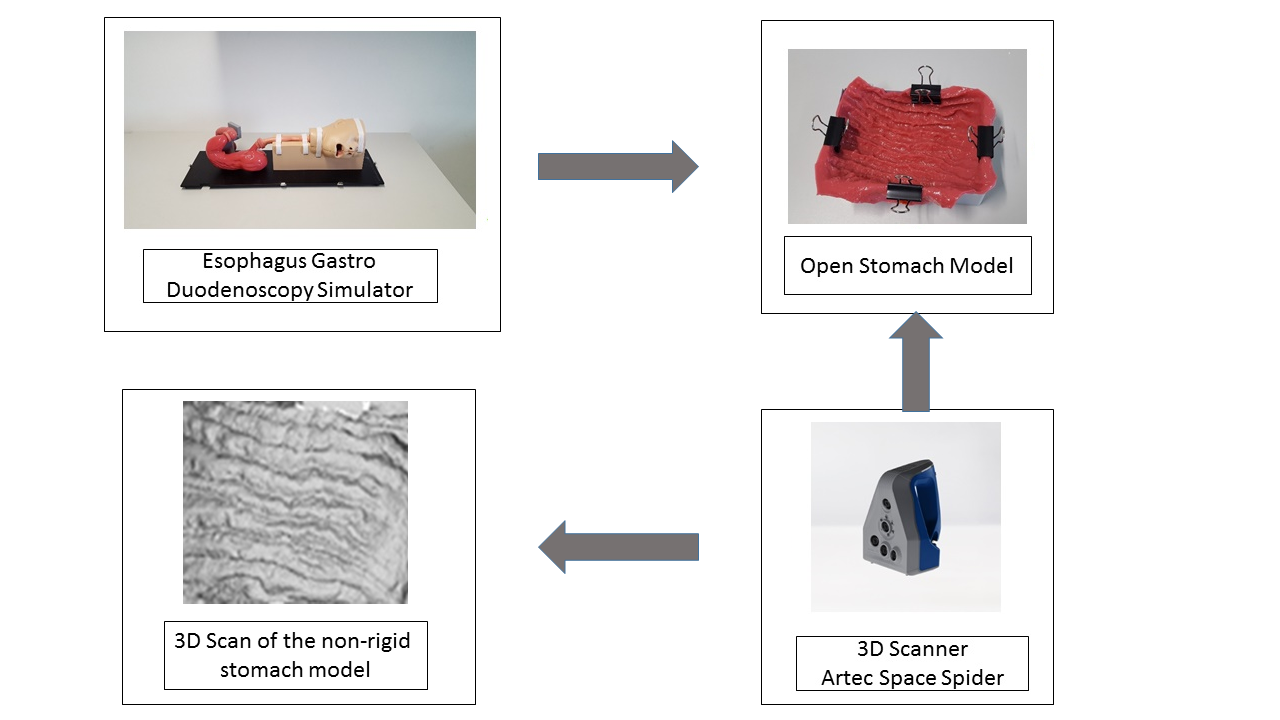}
\caption{3D scan of the open non-rigid stomach model by 3D Artec Space Spider scanner.}
\label{fig:3d_scan}       
\end{figure}

Since there is no publicly available dataset existing for 6-DoF endoscopic capsule robot localization and mapping to our knowledge, we created our authentic capsule robot dataset with ground truths. To make sure that our dataset was not narrowed down to just one specific endoscopic camera, three different endoscopic cameras were used to capture the endoscopic videos. The cameras used for recording the dataset were the AWAIBA Naneye, VOYAGER and POTENSIC endoscopic camera. Specifications of the cameras can be found in Table \ref{tab:cam_1}, Table \ref{tab:cam_2} and Table  \ref{tab:cam_3}, respectively. We mounted  endoscopic cameras on our magnetically activated soft capsule endoscope  (MASCE) systems as seen in Figure \ref{fig:masce}. The videos were recorded on an oiled non-rigid, realistic surgical stomach model Koken LM103 - EDG (EsophagoGastroDuodenoscopy) Simulator. 
To obtain 6-DoF localization ground truth for the captured videos, OptiTrack motion tracking system consisting of four infrared cameras and a tracking software was utilized. Figure \ref{fig:ex} demonstrates our experimental setup as a visual reference. A total of 15 minutes of stomach imagery was recorded containing over 10000 frames. Some sample frames of the dataset are shown in Figure \ref{fig:datas}. Finally, we scanned the open surgical stomach model using the 3D Artec Space Spider image scanner. This 3D image scan served as the ground truth for the error calculation for our 3D map reconstruction system (\ref{fig:3d_scan}).

\begin{table}
\parbox{.45\linewidth}{
\caption{AWAIBA NANEYE MONOCULAR CAMERA SPECIFICATIONS}
\label{tab:cam_1}       
\begin{tabular}{ll}
\hline\noalign{\smallskip}
Resolution & 250 x 250 pixels \\
\hline\noalign{\smallskip}
Footprint & 2.2 x 1.0 x 1.7 mm \\
\hline\noalign{\smallskip}
Pixel size & 3 x 3 $\mu m^2$\\
\hline\noalign{\smallskip}
Pixel depth & 10 bit\\
\hline\noalign{\smallskip}
Frame rate & 44 fps\\
\hline\noalign{\smallskip}
\end{tabular}
}
\hfill
\parbox{.45\linewidth}{
\caption{POTENSIC MINI MONOCULAR CAMERA SPECIFICATIONS}
\label{tab:cam_2}       
\begin{tabular}{ll}
\hline\noalign{\smallskip}
Resolution & 1280 x 720 pixels \\
\hline\noalign{\smallskip}
Footprint & 5.2 x 4.0 x 2.7 mm \\
\hline\noalign{\smallskip}
Pixel size & 10 x 10 $\mu m^2$\\
\hline\noalign{\smallskip}
Pixel depth & 10 bit\\
\hline\noalign{\smallskip}
Frame rate & 15 fps\\
\hline\noalign{\smallskip}
\end{tabular}
}
\end{table}

\begin{table}
\centering
\caption{VOYAGER MINI CAMERA SPECIFICATIONS}
\label{tab:cam_3}       
\begin{tabular}{ll}
\hline\noalign{\smallskip}
Resolution & 720 x 480 pixels \\
\hline\noalign{\smallskip}
Footprint & 5.2 x 5.0 x 2.7 mm \\
\hline\noalign{\smallskip}
Pixel size & 10 x 10 $\mu m^2$\\
\hline\noalign{\smallskip}
Pixel depth & 10 bit\\
\hline\noalign{\smallskip}
Frame rate & 15 fps\\
\hline\noalign{\smallskip}
\end{tabular}
\end{table}


\subsection{Trajectory Estimation} \label{sec:trajectory_estimation}

\begin{figure}[t!]
\begin{subfigure}[t]{\textwidth} 
\includegraphics[width=\textwidth]{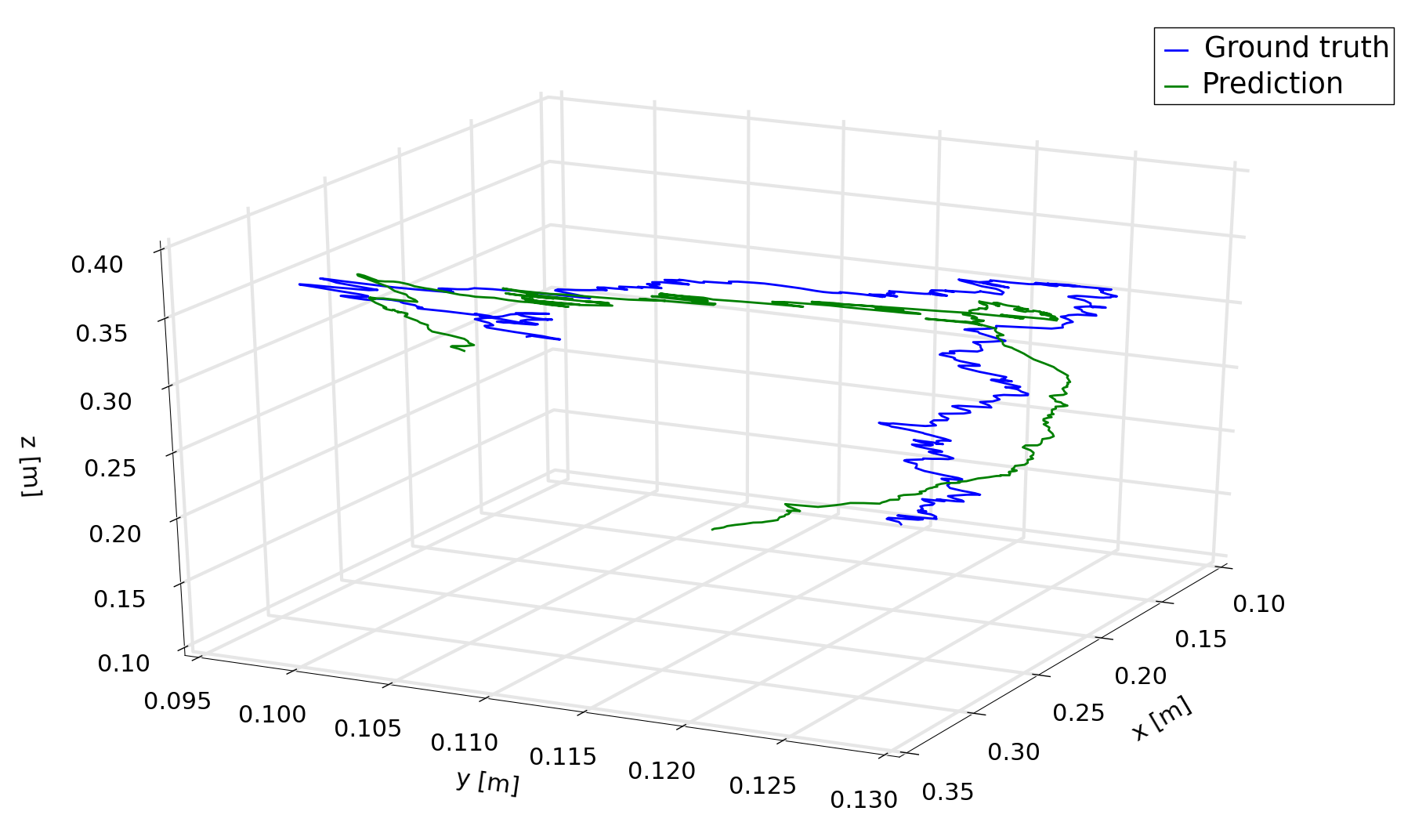}
\caption{Tracked robot trajectory vs Ground truth 1 }
\label{fig:cross_trans}       
\end{subfigure}
\\
\begin{subfigure}[t]{\textwidth} 
\includegraphics[width=\textwidth]{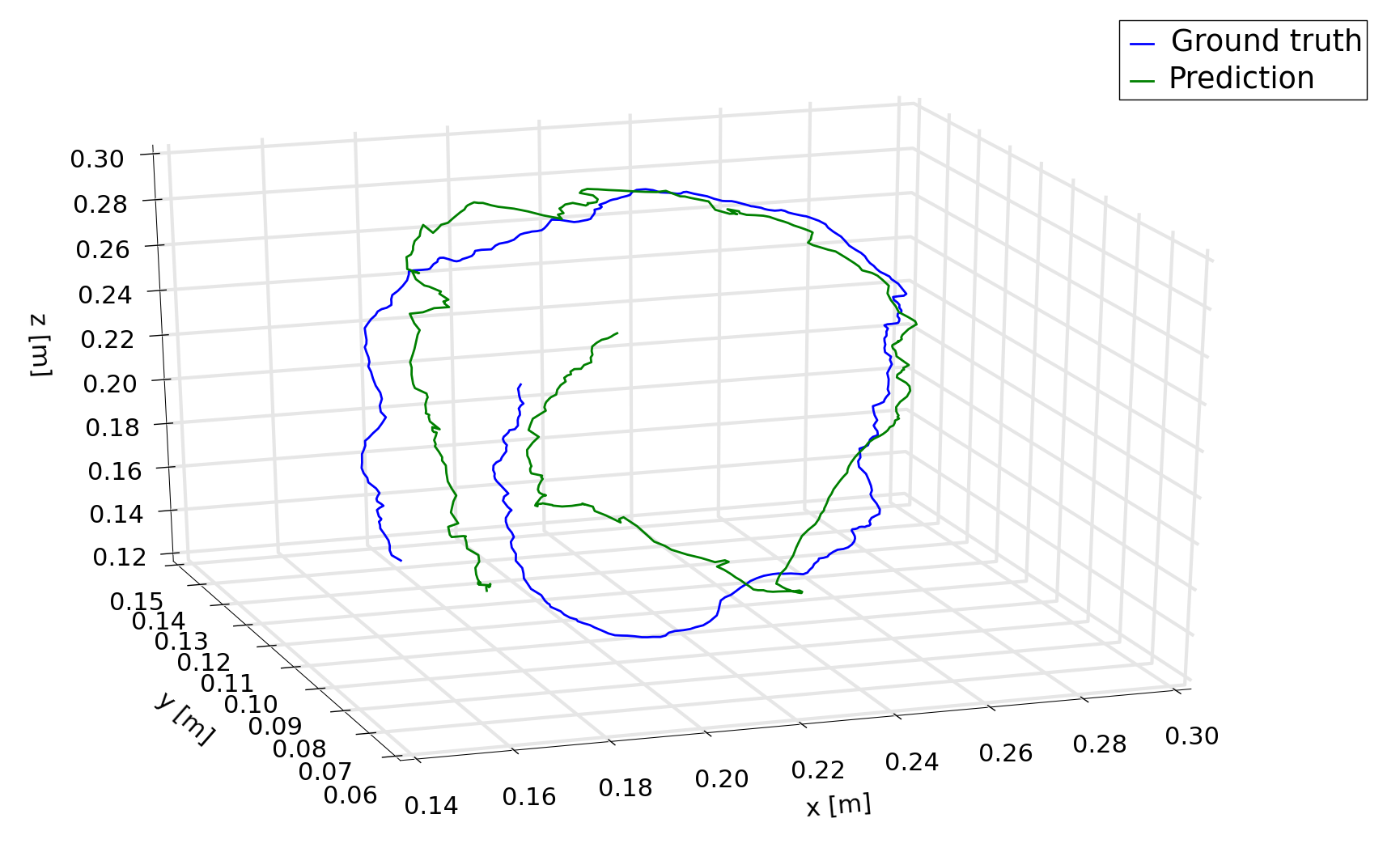}
\caption{Tracked robot trajectory vs Ground truth 2 }
\label{fig:cross_rot} 
\end{subfigure}
\caption{ }
\label{fig:cross_val}
\end{figure}

\begin{figure}[t!]
\begin{subfigure}[t]{\textwidth} 
\includegraphics[width=\textwidth]{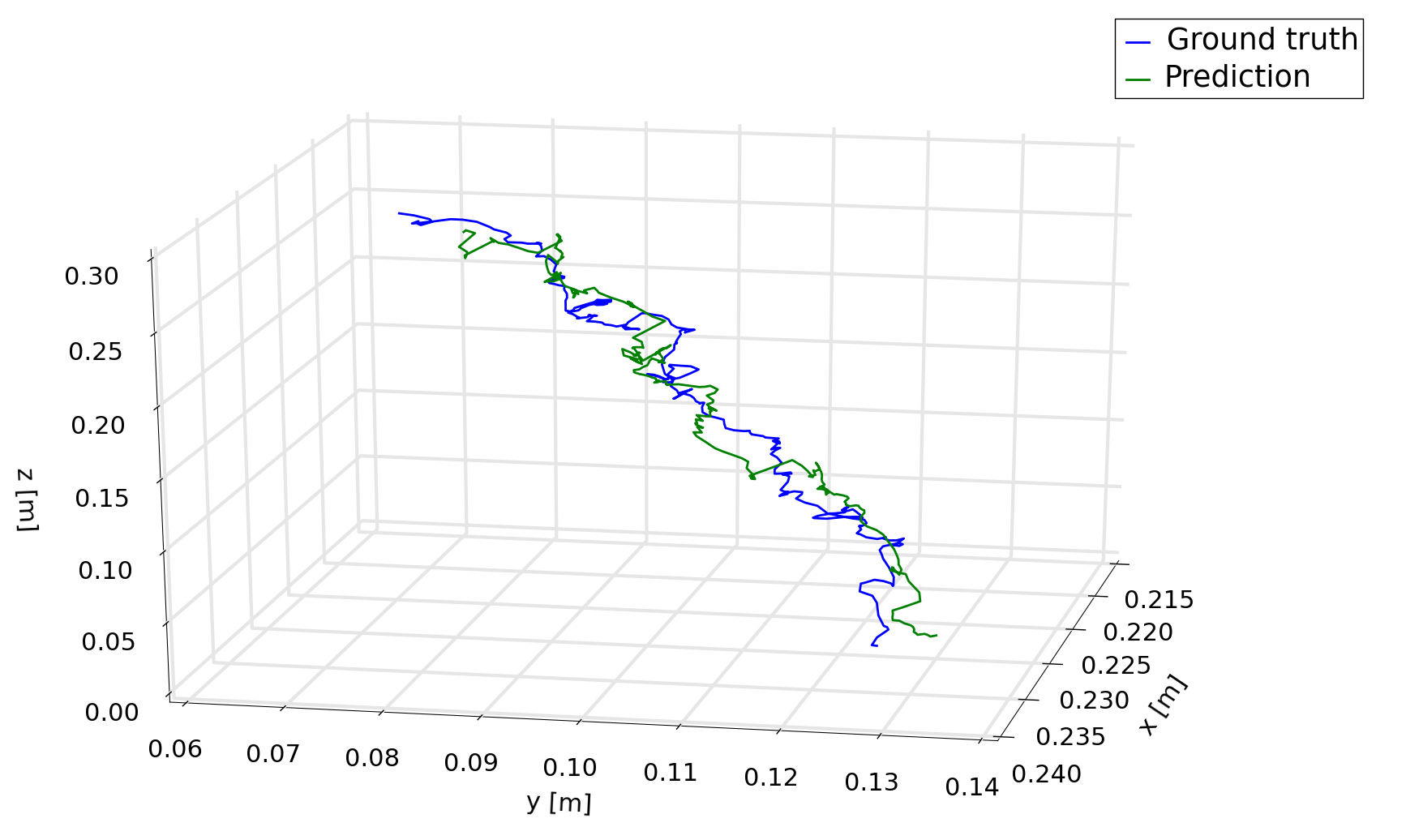}
\caption{Tracked robot trajectory vs Ground truth 3}
\label{fig:cross_rotq} 
\end{subfigure}
\\
\begin{subfigure}[t]{\textwidth} 
\includegraphics[width=\textwidth]{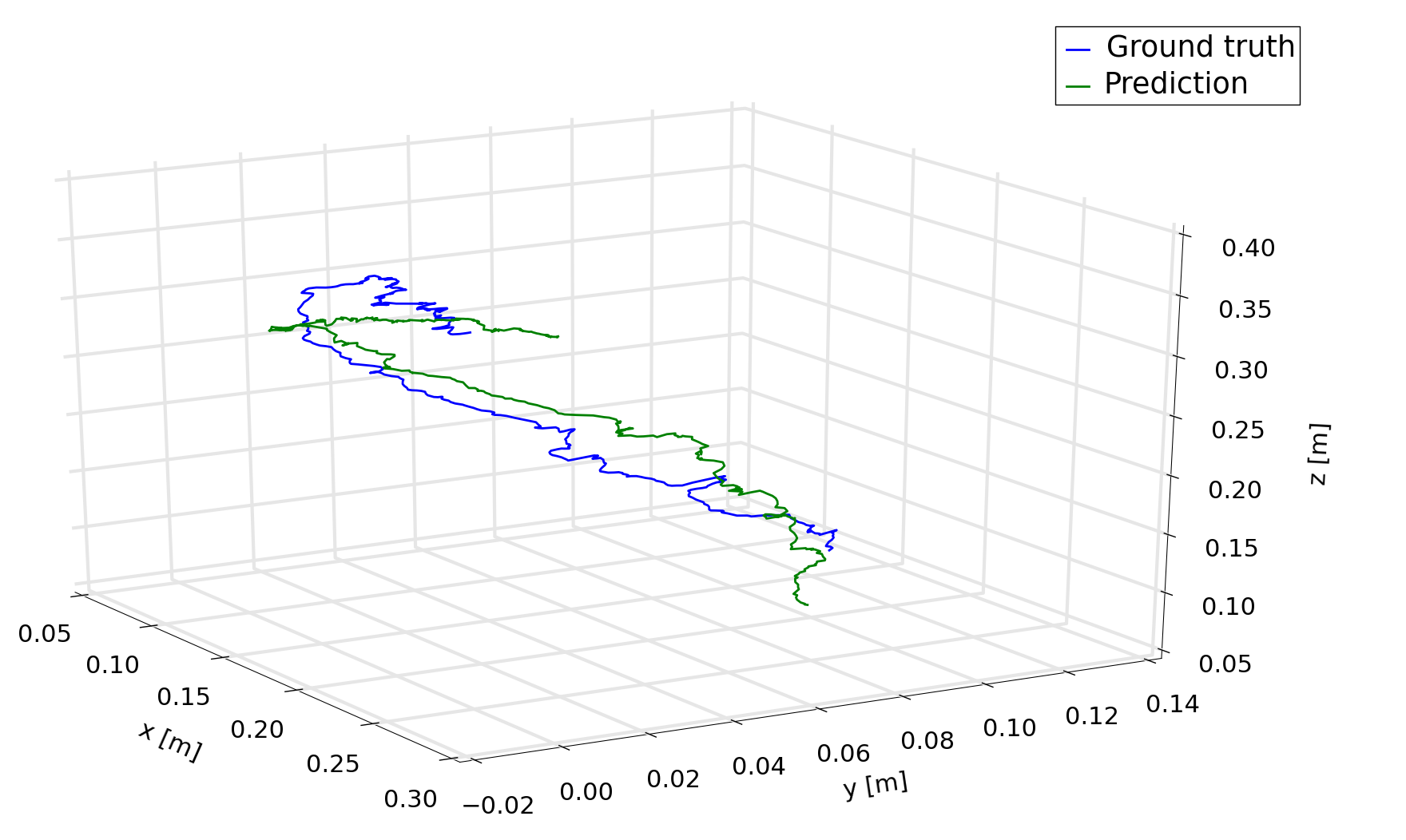}
\caption{Tracked robot trajectory vs Ground truth 4}
\label{fig:cross_rotw} 
\end{subfigure}
\caption{ }
\label{fig:cross_vale}
\end{figure}

Table \ref{tab:trans_rmse} demonstrates the results of the trajectory estimation for 7 different trajectories. Trajectory 1 is an uncomplicated path with slow incremental translations and rotations. Trajectory 2 follows a comprehensive scan of the stomach with many local loop closures. Trajectory 3 contains an extensive scan of the stomach with more complicated local loop closures. Trajectory 4 consists of more challenge motions including faster rotational and translational movements. Trajectory 5 consists of very loopy and complex motions with many loop closures. Trajectory 6 is the same of trajectory 5 but included synthetic noise to see the robustness of the system against noise effects. Before capturing trajectory 7, we added more paraffin oil into the simulator tissue to have heavier reflection conditions. Similar to the trajectory 6, trajectory 7 consists of very loopy and complex motions including very fast rotations, translations and drifting. Some qualitative tracking results of our proposed system and corresponding ground truth trajectory sections are demonstrated in figures \ref{fig:cross_trans},\ref{fig:cross_rot},\ref{fig:cross_rotq}, and \ref{fig:cross_rotw}. For the quantitative analysis, we used the absolute trajectory (ATE) root-mean-square error metric (RMSE), which measures the root-mean-square of the euclidean distances between all estimated camera poses and the ground truth poses associated by timestamp \citep{sturm2012benchmark}. As seen in table \ref{tab:trans_rmse}, the system performs a very robust and accurate tracking in all of the challenge datasets and is not affected by sudden movements, blur, noise or heavy spectral reflections.

\begin{table}
	\centering
	\caption{Trajectory lengths and RMSE results in meters}
	\label{tab:trans_rmse}       
	\begin{tabular}{ccccc}
		\hline\noalign{\smallskip}
		 Trajectory ID & POTENSIC & VOYAGER & AWAIBA& LENGTH \\
		\hline\noalign{\smallskip}
		1&0.015&0.018&0.020& 0.414 \\
		\hline\noalign{\smallskip}
		2&0.018&0.021&0.023& 0.513 \\
		\hline\noalign{\smallskip}
		3&0.017&0.019&0.025& 0.432 \\
		\hline\noalign{\smallskip}
		4&0.032&0.035&0.042& 0.478 \\
		\hline\noalign{\smallskip}
		5&0.035&0.038&0.045& 0.462 \\
		\hline\noalign{\smallskip}
		6&0.038&0.041&0.048& 0.481 \\
		\hline\noalign{\smallskip}
		7&0.041&0.043&0.049& 0.468 \\
		\hline\noalign{\smallskip}
	\end{tabular}
\end{table}

\subsection{Surface Estimation} \label{sec:surface_estimation}

\begin{table}
	\centering
	\caption{Trajectory length and 3D surface reconstruction RMSE results in meters}
	\label{tab:3drmse}       
	\begin{tabular}{ccccc}
		\hline\noalign{\smallskip}
		Trajectory ID  & POTENSIC & VOYAGER & AWAIBA& Length \\
		\hline\noalign{\smallskip}
		1 & 0.023 &0.025&0.028& 0.414 \\
		\hline\noalign{\smallskip}
		2 & 0.025 &0.027&0.032& 0.513 \\
		\hline\noalign{\smallskip}
		3 & 0.026 &0.029&0.034& 0.432 \\
		\hline\noalign{\smallskip}
		4 & 0.029 &0.032&0.035& 0.478 \\
		\hline\noalign{\smallskip}
		5 & 0.032 &0.034&0.038& 0.462 \\
		\hline\noalign{\smallskip}
		6 & 0.034 &0.036&0.041&0.481 \\
		\hline\noalign{\smallskip}
		7 & 0.035 &0.041&0.044& 0.468 \\
		\hline\noalign{\smallskip}
	\end{tabular}
\end{table}
\begin{figure}
\centering
  \includegraphics[width=\textwidth]{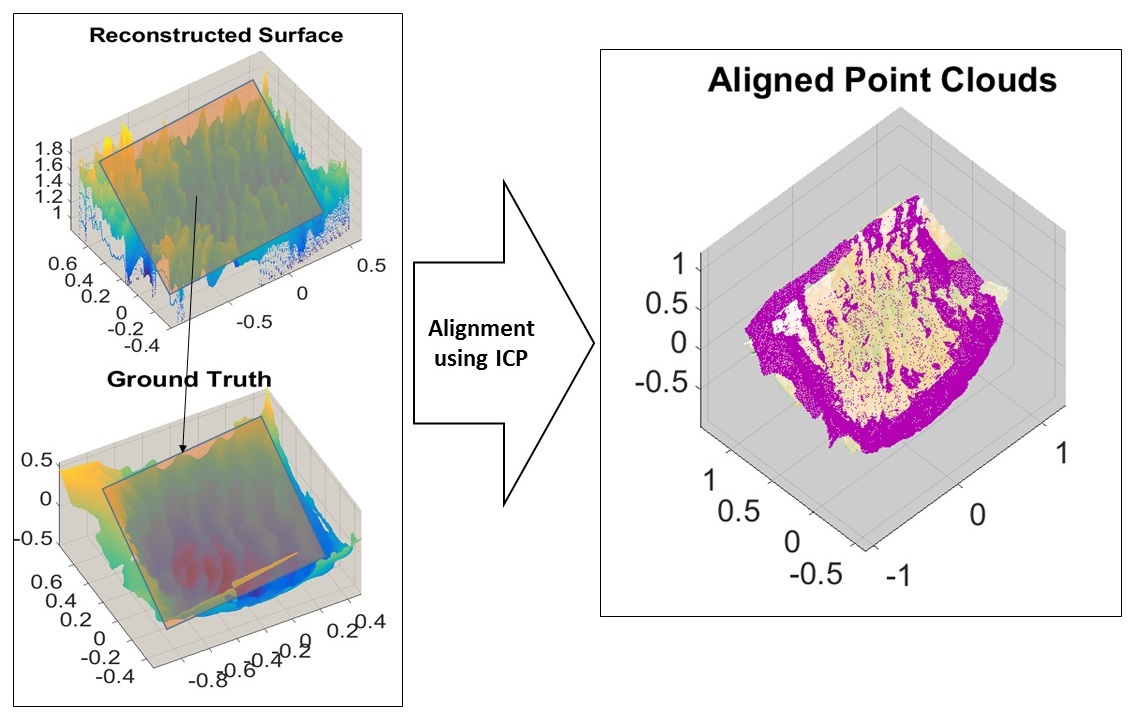}
\caption{Alignment of reconstructed surface and ground truth.}
\label{fig:pointcloud_alignment}       
\end{figure}

\begin{figure}
\centering
  \includegraphics[width=\textwidth]{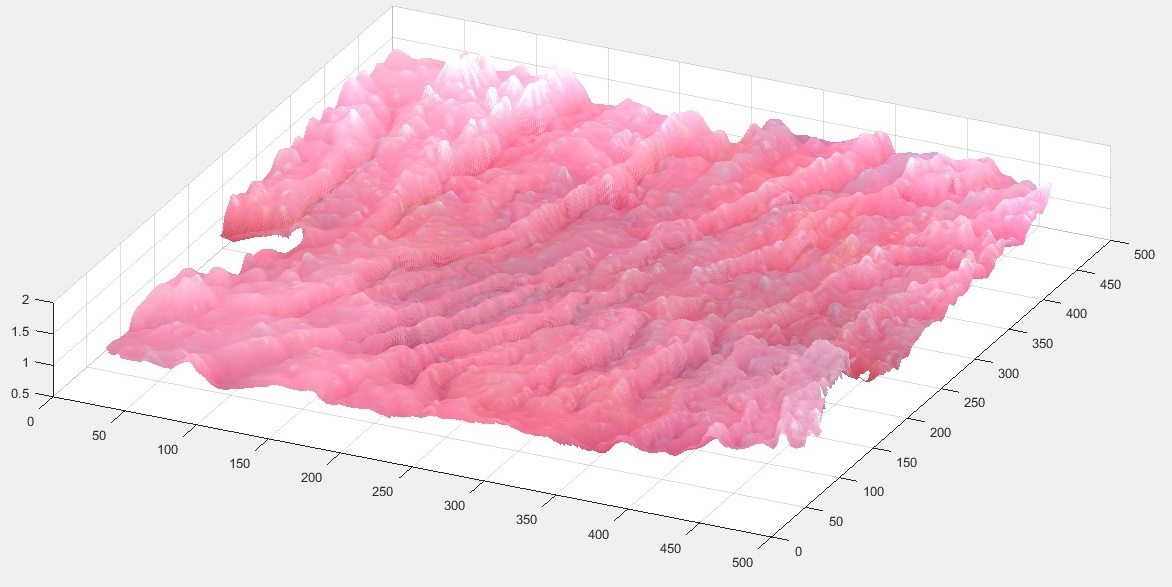}
\caption{3D reconstructed soft stomach simulator surface}
\label{fig:recounstrudted_map}       
\end{figure}

We evaluated the surface reconstruction results of our system on the same dataset we used for the trajectory estimation as well. We scanned the non-rigid EGD (Esophagogastroduodenoscopy) simulator to obtain the ground truth 3D data. Reconstructed 3D surface and ground truth 3D data were aligned using iterative closest point algorithm (ICP). RMSE for reconstructed surface was calculated using again the absolute trajectory (ATE) RMSE, which measured this time the root-mean-square of the euclidean distances between estimated depth values and the corresponding ground truth depth values \citep{sturm2012benchmark} (Figure \ref{fig:pointcloud_alignment}). The detailed RMSE results in Table \ref{tab:3drmse} prove that even in very challenge trajectories such as trajectory 4, 5, 6, and 7 with sudden movements, heavy noise and reflections, our system is capable of providing a reliable and accurate 3D surface reconstruction of the explored inner organ tissue. A sample 3D reconstructed surface is shown in Figure \ref{fig:recounstrudted_map} for visual reference.

\subsection{Computational Performance} \label{sec:computational_performance} 

To analyze the computational performance of the system we observed the average frame processing time across the trajectory 2 and 4 sequences. The test platform was a desktop PC with an Intel Xeon E5-1660v3- CPU at 3.00, 8 cores, 32GB of RAM and an NVIDIA Quadro K1200 GPU with 4GB of memory. The execution time of the system is depended on the number of surfels in the map, with an overall average of 48 ms per frame scaling to a peak average of 53 ms implying a worst case processing frequency of 18 Hz.

\section{CONCLUSION} \label{sec:conclusion}

Endoscopic capsule robots are promising novel, minimally invasive  technological developments in the area of medical devices for the GI tract. In this paper, we have presented for the first time in the literature a visual SLAM method for endoscopic capsule robots. Our system makes use of shape from shading surface reconstruction technique to obtain the depth sense of the consecutive frames and performs time windowed surfel-based dense data fusion in combination with frame-to-model tracking and non-rigid deformation. The proposed system was able to produce a highly accurate 3D map of the explored inner organ tissue and was able to stay close to the ground truth endoscopic capsule robot trajectory even for very challenge datasets confronted with sharp rotations and fast translations, heavy specular reflections and noise. Our system proved qualitatively and quantitatively its effectiveness in occasionally looping capsule robot motions and comprehensive inner organ scanning tasks. In future, we aim to extend our work into the stereo capsule endoscopy applications to achieve even more accurate localization and mapping and demonstrate the accuracy of our results in animal experiments.

\section*{References}

\bibliography{mybibfile}

\end{document}